\documentclass[sigconf]{acmart}

\usepackage{booktabs}
\usepackage{multirow}
\usepackage{xcolor}
\usepackage{adjustbox}
\usepackage{threeparttable}
\usepackage{bm}
\usepackage{mathrsfs}

\AtBeginDocument{%
  \providecommand\BibTeX{{%
    \normalfont B\kern-0.5em{\scshape i\kern-0.25em b}\kern-0.8em\TeX}}}

\copyrightyear{2020} 
\acmYear{2020} 
\setcopyright{acmcopyright}\acmConference[MM '20]{Proceedings of the 28th ACM International Conference on Multimedia}{October 12--16, 2020}{Seattle, WA, USA}
\acmBooktitle{Proceedings of the 28th ACM International Conference on Multimedia (MM '20), October 12--16, 2020, Seattle, WA, USA}
\acmPrice{15.00}
\acmDOI{10.1145/3394171.3413908}
\acmISBN{978-1-4503-7988-5/20/10}

\begin{document}
\fancyhead{}

\title{Controllable Video Captioning with an Exemplar Sentence}

\author{Yitian Yuan}
\authornote{This work was done while Yitian Yuan was a Research Intern at Tencent AI Lab.}
\email{yyt18@mails.tsinghua.edu.cn}
\affiliation{
  \institution{Tsinghua-Berkeley Shenzhen Institute,\\Tsinghua University}
}

\author{Lin Ma$^\dagger$}
\email{forest.linma@gmail.com}
\affiliation{
  \institution{Meituan}
}

\author{Jingwen Wang}
\email{jaywongjaywong@gmail.com}
\affiliation{
  \institution{Tencent AI Lab}
}

\author{Wenwu Zhu}
\authornote{Corresponding authors.}
\email{wwzhu@tsinghua.edu.cn}
\affiliation{
  \institution{Department of Computer Science and Technology, Tsinghua University}
}

\renewcommand{\shortauthors}{Yuan, et al.}

\begin{abstract}
In this paper, we investigate a novel and challenging task, namely controllable video captioning with an exemplar sentence. Formally, given a video and a syntactically valid exemplar sentence, the task aims to generate one caption which not only describes the semantic contents of the video, but also follows the syntactic form of the given exemplar sentence. In order to tackle such an exemplar-based video captioning task, we propose a novel Syntax Modulated Caption Generator (SMCG) incorporated in an encoder-decoder-reconstructor architecture. The proposed SMCG takes video semantic representation as an input, and conditionally modulates the gates and cells of long short-term memory network with respect to the encoded syntactic information of the given exemplar sentence. Therefore, SMCG is able to control the states for word prediction and achieve the syntax customized caption generation. We conduct experiments by collecting auxiliary exemplar sentences for two public video captioning datasets. Extensive experimental results demonstrate the effectiveness of our approach on generating syntax controllable and semantic preserved video captions. By providing different exemplar sentences, our approach is capable of producing different captions with various syntactic structures, thus indicating a promising way to strengthen the diversity of video captioning. Code for this paper is available at \texttt{\url{https://github.com/yytzsy/SMCG}}.
\end{abstract}


\maketitle

\section{Introduction}

\begin{figure}[!t]
\centering
\setlength{\abovecaptionskip}{0.cm}
\setlength{\belowcaptionskip}{-0.cm}
\includegraphics[width=3.3in]{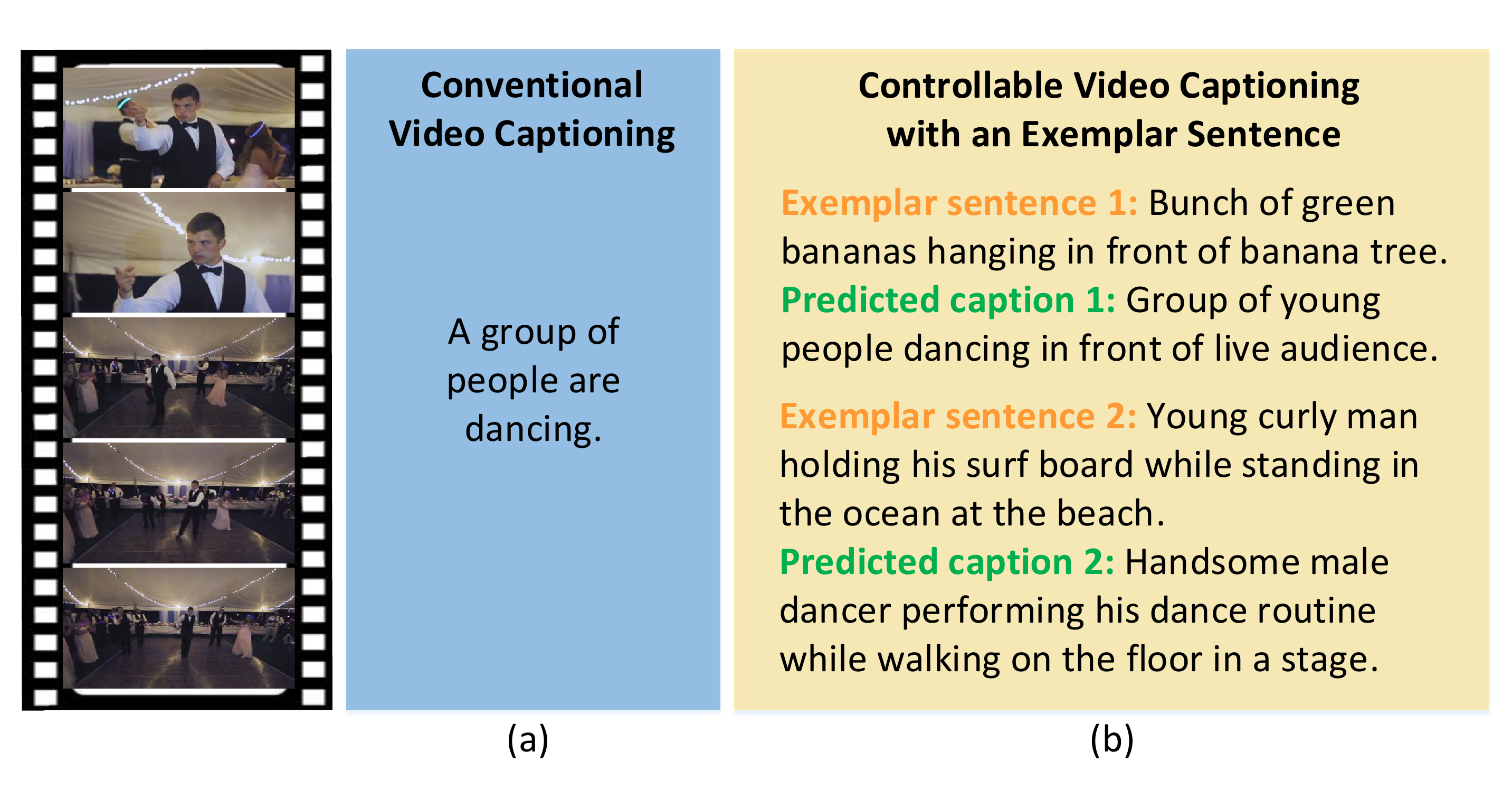}
\caption{ The comparison between the conventional video captioning task (a) and our proposed controllable video captioning with an exemplar sentence task (b).}
\label{fig:intro}
\vspace{-0.2in}
\end{figure}

Automatically generating sentence descriptions for videos, i.e., video captioning, has emerged as a prominent research problem to bridge computer vision and natural language processing. From early template-based approaches \cite{kojima2002natural,guadarrama2013youtube2text,rohrbach2013translating,rohrbach2014coherent,xu2015jointly} to recent sequence learning approaches \cite{venugopalan2015sequence,venugopalan2014translating,pan2016jointly,yao2015describing,pan2017video,chen2018less,wang2018reconstruction,yu2016video}, remarkable advancements have been achieved on this task. However, most conventional video captioning models mainly focus on understanding video semantics so as to produce captions describing video contents, while the linguistic expressiveness and diversity of the video captions are often ignored. For example, as shown in Figure \ref{fig:intro}(a), by learning from the caption annotations with simple linguistic characteristics in the training set, the predicted caption is monotonous and plain, which will further influence the user experiences of potential practical applications, such as chat robots \cite{Kottur_2018_ECCV,
visdial_rl}, automatic video commenting \cite{li2016video}, etc.

To enrich the linguistic expressiveness of visual captions, some researchers~\cite{Gan2017StyleNet,Mathews2018SemStyle,YouImage} incorporate specific styles, such as humorous or romantic, into the captioning process. Despite the promising results, such stylized video captioning models still constrain the generated captions in a predefined set of linguistic styles, and may limit the diversity and flexibility of video captions. Recently, the part-of-speech (POS) tag sequences are further introduced to control the syntactic structure of the predicted captions~\cite{deshpande2019fast,wang2019controllable}. However, directly manipulating the POS tag needs specialized knowledge on the sentence syntax and grammar rules, which is hard to be realized practically.

Compared with style labels or POS sequences, one intuitive and straightforward way is to directly leverage one exemplar sentence to control or customize the video caption generation. Therefore, it motivates us to propose and investigate a novel task in this paper, namely controllable video captioning with an exemplar sentence or exemplar-based video captioning. As shown in Figure \ref{fig:intro}(b), given any syntactically valid exemplar sentence and a video, the task aims to generate one caption, which can not only express the semantic contents of the video, but also follow the syntactic structure of the given exemplar sentence. Since exemplar sentences are easy to acquire and have no specific constraint, this task can generate diverse captions with a variety of syntactic structures, thus strengthening the expressiveness and diversity of the video captions. 

The proposed exemplar-based video captioning is a challenging problem. Firstly, how to extract syntactic information from exemplar sentences, and make generated captions absorb the extracted sentence syntaxes is quite crucial. Secondly, since introducing extra exemplar sentences will also bring some disturbances for semantic contents of the input video, how to preserve video semantics in the generated captions is worthy of further considerations. 

\begin{figure*}[!t]
\centering
\setlength{\abovecaptionskip}{0.cm}
\setlength{\belowcaptionskip}{-0.cm}
\includegraphics[width=0.85\textwidth]{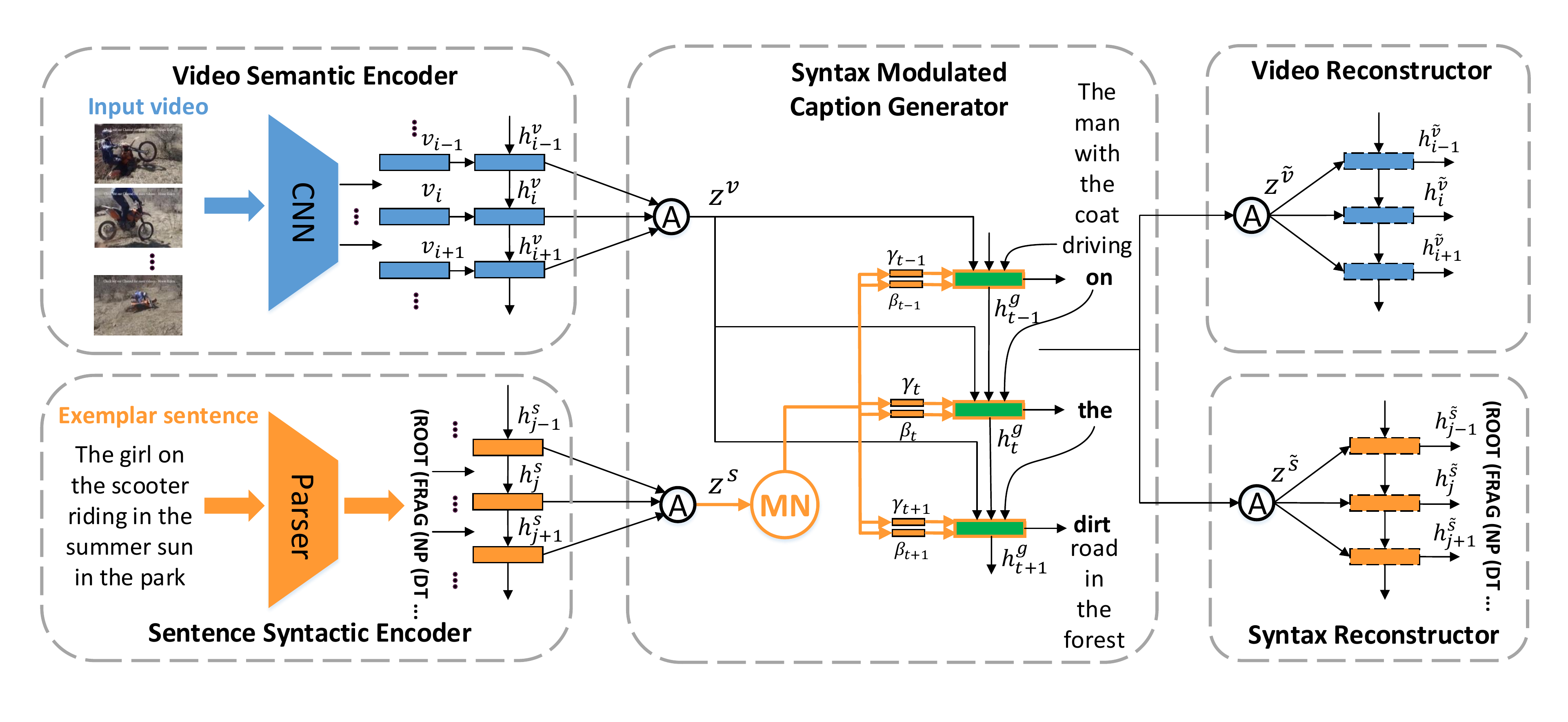}
\caption{ An overview of our proposed model, which is realized in an encoder-decoder-reconstructor architecture. The video semantic and sentence syntactic encoders extract visual semantic and sentence syntactic information from the input video and the given exemplar sentence, respectively. The syntax modulated caption generator attentively summarizes the  video semantic information and relies on the syntactic information to modulate the semantic word decoding procedure for the exemplar-based caption generation. To further  preserve video semantic and exemplar sentence syntactic information, the video and syntax reconstructors reproduce the original video features and exemplar syntax tokens from the generated caption, respectively. Here \textcircled{\tiny{MN}} denotes our proposed modulation network, and \textcircled{\tiny{A}} denotes the soft attention mechanism.}
\label{fig:framework}
\end{figure*}

In order to tackle the aforementioned challenges, we propose a novel Syntax Modulated Caption Generator (SMCG) incorporated in one encoder-decoder-reconstructor architecture. As shown in Figure \ref{fig:framework}, the constituency parse tree of the exemplar sentence is firstly extracted by a sentence parser. Afterwards, two encoders then process the exemplar sentence parse tree and video feature sequence to obtain the syntactic and semantic representations, respectively. The proposed SMCG, which summarizes the semantic representation as input and utilizes the syntactic representation to modulate the gates and cells in the sentence decoding long short-term memory (LSTM)~\cite{hochreiter1997long}, is thereby able to generate one caption expressing the semantic meaning of the video and meanwhile following the syntactic structure of the exemplar sentence. Moreover, two reconstructors built on the hidden states of the caption generator are established to reproduce the original video feature and exemplar sentence parse tree, respectively. Such reconstructions can further help preserve video semantics in predicted captions and ensure the syntax customization.



To summarize, the contributions of this work lie in three folds:
\begin{itemize}
    \item We propose a novel task --- controllable video captioning with an exemplar sentence. Provided with widely accessible and various exemplar sentences, it can strengthen the diversity and expressiveness of video captions in an intuitive and natural way.
    \item A novel Syntax Modulated Caption Generator (SMCG) is proposed, which relies on the syntactic information of the exemplar sentence to modulate the sequential word decoding procedure, and thus controls the syntactic structures of the predicted sentence while preserving the semantic meanings of the original video contents.
    \item We conduct the exemplar-based video captioning experiments by collecting auxiliary exemplar sentences for two public video datasets. Extensive experimental results demonstrate that our model can generate captions which not only precisely describe video contents, but also possess similar syntactic structures to the exemplar sentences.
\end{itemize}

\section{Related Work}

\subsection{Video Captioning}
Early works \cite{kojima2002natural,guadarrama2013youtube2text,rohrbach2013translating,rohrbach2014coherent,wang2020reconstruct,xu2015jointly,wang2018bidirectional} on video captioning adopted template-based methods, which first define some specific rules for language grammar, and then generate captions by associating words detected from visual contents with predefined sentence templates. Currently, with the development of deep neural networks, sequence learning methods with an encoder-decoder architecture \cite{venugopalan2014translating} have been widely adopted for video captioning. Based on such an architecture, numerous improvements were introduced, such as sequential encoding of video features \cite{venugopalan2015sequence}, soft attention \cite{yao2015describing}, visual-semantic embedding \cite{pan2016jointly}, dual learning \cite{wang2018reconstruction}, hierarchical recurrent neural network \cite{yu2016video,baraldi2017hierarchical}, multimodal feature fusion \cite{xu2017learning}, multi-task learning \cite{pasunuru2017multi}, reinforcement learning \cite{pasunuru2017reinforced,wang2018video}, etc.

Although promising results have been achieved, template-based methods constrain the video captions with fixed syntactic structures, while sequence learning methods generate sentence descriptions by fitting the monotonous linguistic patterns of captions from the training set. All of those methods ignore that one single video can be described in a variety of sentences with diverse expressions. In this paper, we propose a problem of controllable video captioning with an exemplar sentence, which is expected to control the syntactic structures of the predicted captions by imitating different exemplar sentences, and thus strengthen the expressiveness and diversity of video captions.

\subsection{Controllable Captioning with Auxiliary Information Guidance}
Recently, some research works start to utilize auxiliary information to control the captioning procedure. A part of works focused on generating captions satisfying certain stylistic requirements such as being humorous or exhibiting a particular sentiment \cite{Gan2017StyleNet,Mathews2018SemStyle,YouImage}. Hence, style indicators or labels were introduced to guide the caption generator. However, those works still typically assume a known, finite set of values that the style attribute can take on, which may still limit the diversity and flexibility of caption generation. Besides, to make the generated captions more diverse and accurate, Deshpande \textit{et al.} leveraged the quantized Part-of-Speech (POS) tag sequence sampled from a given benchmark to condition word prediction at the decoding recurrent model \cite{deshpande2019fast}. Wang \textit{et al.} tried to predict the POS sequence tag by tag from the input video, and then embeded them as a global POS representation to gate the inputs of the sentence decoder for syntax control \cite{wang2019controllable}. With manually altering the predicted POS tag sequence, Wang \textit{et al.} showed that they can obtain captions with different syntaxes. However, controlling caption generation by manually manipulating POS tag needs specialized linguistic knowledge and is not intuitive, which can hardly be applied in practical scenarios.

Our proposed controllable video captioning with an exemplar sentence task is different from the above works. The syntactic guidance for video captioning is derived from the given exemplar sentences, which are easy to access and have various syntactic structures. Thus, it can naturally extend the diversity of video captioning. Moreover, controlling caption generation with an exemplar sentence is more intuitive for people to realize and evaluate.

\section{The Proposed Model}



In this paper, we propose a novel model to solve the controllable video captioning with an exemplar sentence task. As shown in Figure \ref{fig:framework}, our proposed model, which is realized in an encoder-decoder-reconstructor architecture, consists of a pair of semantic and syntactic encoders, a syntax modulated caption generator, and a pair of video and syntax reconstructors. Please note that the whole architecture is fully coupled together and can therefore be trained in an end-to-end manner.


\subsection{Semantic and Syntactic Encoders }

\noindent \textbf{Video Semantic Encoder.} Given an input video, we first use one pretrained CNN to encode each video frame into a fixed-length representation.
In this way, the given video is encoded as a sequential representation $\mathbf{V} = \{\mathbf{v}_1,\mathbf{v}_2,\cdots,\mathbf{v}_m\}$, where $m$ denotes the total number of the video frames. To incorporate the context information, an LSTM is used to aggregate the sequential representation $\mathbf{V}$ as follows:
\begin{equation}
\small
\setlength{\abovedisplayskip}{3pt}
\setlength{\belowdisplayskip}{3pt}
    \{\mathbf{h}_i^{v}, \mathbf{c}_i^{v}\} = {\rm LSTM}_{v}^{E} (\mathbf{v}_i,\mathbf{h}_{i-1}^{v}, \mathbf{c}_{i-1}^{v}).
    \label{encoder_video}
\end{equation}
Hence, we obtain the context incorporated video sequence representation  $\mathbf{H}^{v} = \{\mathbf{h}_1^{v},\mathbf{h}_2^{v},\cdots,\mathbf{h}_m^{v}\}$, with $\mathbf{H}^{v}$ encoding the semantic information of the video sequence.

\noindent \textbf{Sentence Syntactic Encoder.} To syntactically control the target caption generation, we first use a sentence parser to extract the constituency parse tree~\cite{Manning2014The,shen2018straight,kitaev2018constituency} of the given exemplar sentence. For the convenience of further processing, the leaf nodes (i.e., word tokens) are removed from the extracted parse tree. For example, as shown in Figure \ref{fig:framework}, the obtained parse tree for the exemplar sentence ``\texttt{\small{The girl on the scooter riding in the summer sun in the park}}'' is ``\texttt{\small{(ROOT (FRAG (NP (DT) (NN)) (PP (IN) (NP (NP (DT) (NN)) (VP (VBG) (PP (IN) (NP (NP (DT) (NN) (NN)) (PP (IN)(NP (DT) (NN))))))))))}}''.

We regard the parse tree of the exemplar sentence as a syntactic sequence $S = \{s_1,s_2,\cdots,s_n\}$, where each element, such as ``\texttt{\small{ROOT}}'', ``\texttt{\small{NP}}'', and the bracket ``\texttt{\small{(}}'' or ``\texttt{\small{)}}'' is taken as an independent syntax token in $S$. Another LSTM is then introduced to encode the exemplar syntactic sequence:
\begin{equation}
\small
\setlength{\abovedisplayskip}{3pt}
\setlength{\belowdisplayskip}{3pt}
    \{\mathbf{h}_j^{s}, \mathbf{c}_j^{s}\} = {\rm LSTM}_{s}^{E} (\mathbf{s}_j,\mathbf{h}_{j-1}^{s}, \mathbf{c}_{j-1}^{s}),
     \label{encoder_syntax}
\end{equation}
where $\mathbf{s}_j$ is the embedding of the $j$-th syntax token.
As such, the syntactic representation $\mathbf{H}^{s} = \{\mathbf{h}_1^{s},\mathbf{h}_2^{s},\cdots,\mathbf{h}_n^{s}\}$ is obtained, which will be taken as the syntactic control signal for the further exemplar-based caption generation.

\begin{figure*}[!t]
\centering
\setlength{\abovecaptionskip}{0.cm}
\setlength{\belowcaptionskip}{-0.cm}
\includegraphics[width=0.8\textwidth]{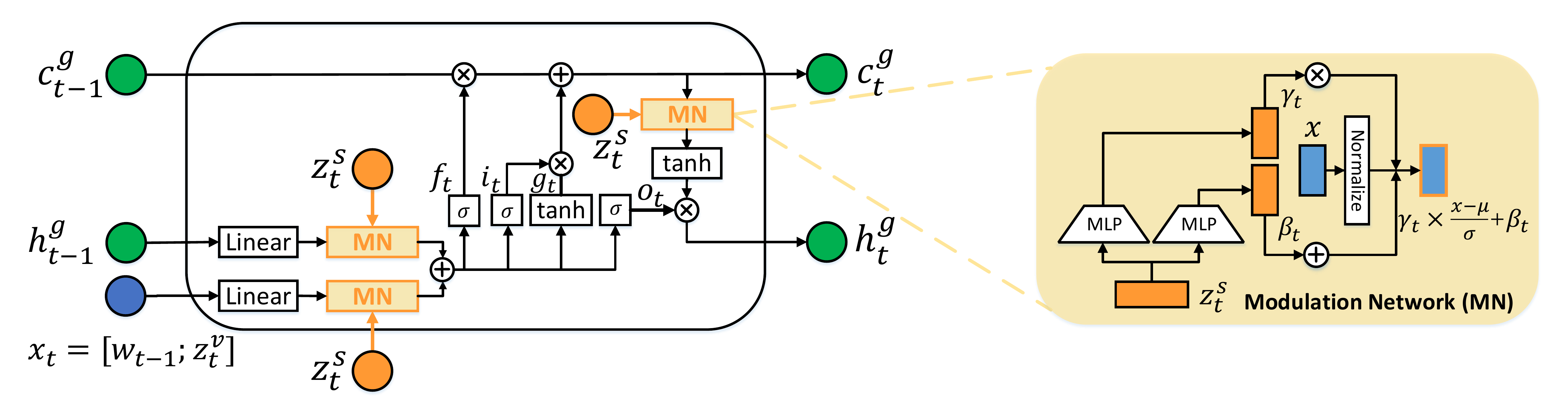}
\caption{ The architecture of our proposed syntax modulated caption generator and the detailed modulation network. Conditioned on the syntactic information, the modulation network manipulates the gate and cell units of the LSTM, and therefore makes the predicted captions imitate the syntactic structure of the exemplar sentence.}
\label{fig:merge_mn}
\vspace{-0.1in}
\end{figure*}

\subsection{Syntax Modulated Caption Generator}

Conventional video captioning models rely on a recurrent neural network (for example, an LSTM) to decode the sentence word by word based on the video semantic inputs. However, for the proposed exemplar-based video captioning task, one important thing is to introduce the exemplar syntactic information into the recurrent model so as to generate syntax customized captions, and meanwhile preserve video semantics.

Inspired by the existing approaches which modulate the normalization parameters of ConvNet under the conditional input guidance for specific tasks (e.g., VQA~\cite{de2017modulating}, temporal sentence grounding~\cite{SCDM19}, and image style transfer~\cite{dumoulin2016learned}), in this work, we propose a novel Syntax Modulated Caption Generator (SMCG) to modulate the gate and cell units of the LSTM with the syntactic cues extracted from one exemplar sentence. The purpose of SMCG is to let the exemplar syntax to manipulate the LSTM gates and cells by scaling them up or down, negating them, or shutting them off \cite{de2017modulating}, so as to control the states of the sentence decoding LSTM as well as the syntactic structures of the decoded sentences. While in this procedure, the video semantic contents are barely modified or mixed with the syntactic information, and are therefore able to be maintained in the predicted captions.


Specifically, suppose that the hidden state of the caption decoding LSTM at timestep $t-1$ is $\mathbf{h}_{t-1}^g$, we first attentively summarize the encoded video semantic and sentence syntactic representations~\cite{yao2015describing} as follows:
\begin{equation}
\small
\begin{split}
\mathbf{z}^{v}_{t} &= {\rm Attention}_{v}(\mathbf{h}_{t-1}^g,\mathbf{H}^v), \\
\mathbf{z}^{s}_{t} &= {\rm Attention}_{s}(\mathbf{h}_{t-1}^g,\mathbf{H}^s).
\end{split}
\end{equation}

As shown in Figure \ref{fig:merge_mn}, the attentively summarized semantic feature $\mathbf{z}^{v}_{t}$ and the embedding $\mathbf{w}_{t-1}$ of the previously predicted word $w_{t-1}$ are concatenated as the input $\mathbf{x}_t = [ \mathbf{w}_{t-1};\mathbf{z}^{v}_{t} ]$ for updating the states of the LSTM. With the proposed syntactic modulation strategy, the computing flow of the sentence decoding LSTM in SMCG is defined as:
\begin{equation}
\small
\begin{split}
\begin{pmatrix}
    \mathbf{f}_t \\
     \mathbf{i}_t \\
     \mathbf{o}_t \\
     \mathbf{g}_t
\end{pmatrix}
 = {\rm \textbf{MN}} & (\mathbf{W}^h  \mathbf{h}^g_{t-1};\bm{\gamma}^h_t,\bm{\beta}^h_t) + {\rm \textbf{MN}}(\mathbf{W}^x \mathbf{x}_{t};\bm{\gamma}^x_t,\bm{\beta}^x_t) + \mathbf{b},  \\
\mathbf{c}_t^g &= \sigma(\mathbf{f}_t) \odot \mathbf{c}^g_{t-1} + \sigma({\mathbf{i}_t}) \odot {\rm tanh}(\mathbf{g}_t), \\
\mathbf{h}_t^{g} &= \sigma(\mathbf{o}_t) \odot {\rm tanh}({\rm \textbf{MN}}(\mathbf{c}_t^g; \bm{\gamma}^c_t,\bm{\beta}^c_t)),
\end{split}
\label{LSTM_modulation}
\end{equation}
where ${\rm \textbf{MN}}(\mathbf{x};\bm{\gamma}_t,\bm{\beta}_t)$ denotes the modulation network, which firstly performs lay normalization of the input vector and then scales and shifts the normalized vector with regard to the attentively summarized syntactic feature $\mathbf{z}_t^s$:
\begin{equation}
\small
\begin{split}
{\rm \textbf{MN}}(\mathbf{x}  ;\bm{\gamma}_t,\bm{\beta}_t) &= \bm{\gamma}_t \cdot \frac{\mathbf{x}-\bm{\mu}(\mathbf{x})}{\bm{\sigma}(\mathbf{x})} + \bm{\beta}_t \\
\bm{\gamma}_t = f_{\gamma}(\mathbf{z}_t^s),  & \quad
\bm{\beta}_t = f_{\beta}(\mathbf{z}_t^s),
\end{split}
\end{equation}
where $\bm{\mu(\mathbf{x})}$ and  $\bm{\sigma(\mathbf{x})}$ denote the mean and standard deviation, respectively. As illustrated in the right part of Figure \ref{fig:merge_mn}, $f_{\gamma}$ and $f_{\beta}$ denote the two independent multi-layer perceptrons (MLPs) with tanh activation, which yield the modulation vectors $\bm{\gamma}$ and $\bm{\beta}$ to control the feature scaling and shifting, respectively. Since there are three pairs of modulation vectors ($\bm{\gamma}_t^h$,$\bm{\beta}_t^h$), ($\bm{\gamma}_t^x$,$\bm{\beta}_t^x$), and ($\bm{\gamma}_t^c$,$\bm{\beta}_t^c$) in Eq.~(\ref{LSTM_modulation}), three pairs of MLPs with independent parameters are used in our syntax modulated caption generator accordingly.


With the updated syntactically modulated hidden state $\mathbf{h}_t^g$, the $t$-th word in the video caption is predicted by:
\begin{equation}
\small
P(w_t|w_{\textless t}, V, S;\theta) = {\rm softmax} (\mathbf{W}^g \mathbf{h}_t^g + \mathbf{b}^g).
\end{equation}

\subsection{Video and Syntax Reconstructors}
The generated syntax customized video captions in our task should not only express the video semantic meanings but also follow the exemplar sentence syntaxes. Therefore, we propose a pair of video and syntax reconstructors stacking on the caption generator, aiming to reconstruct the video features and the syntax tokens from the hidden states of the caption generator, respectively. The proposed reconstructors are expected to bridge the semantic gaps between the generated captions and the video contents, and meanwhile to close the syntactic distances between the captions and the exemplar sentences.


In this paper, the two reconstructors are both realized in an LSTM architecture, and we only take the video reconstructor~\cite{wang2018reconstruction} as an example for demonstration. Supposing that the hidden state of the video reconstruction LSTM at the $(i\text{-}1)$-th timestep is $\mathbf{h}^{\widetilde{v}}_{i-1}$, the key hidden states from the caption generator are firstly attentively summarized as follows:
\begin{equation}
\small
\setlength{\abovedisplayskip}{5pt}
\setlength{\belowdisplayskip}{5pt}
\mathbf{z}^{\widetilde{v}}_i = {\rm Attention}_{\widetilde{v}}(\mathbf{h}_{i-1}^{\widetilde{v}},\mathbf{H}^g),
\end{equation}
where $\mathbf{H}^g = \{\mathbf{h}_1^g, \mathbf{h}_2^g, \cdots\}$ denotes the collected hidden state sequence of the caption generator. The attentively summarized hidden state $\mathbf{z}^{\widetilde{v}}_i$ is then taken as input to reproduce the original video feature $\mathbf{v}_i$ at the $i$-th timestep:
\begin{equation}
\small
\setlength{\abovedisplayskip}{5pt}
\setlength{\belowdisplayskip}{5pt}
    \mathbf{h}_i^{\widetilde{v}}, \mathbf{c}_i^{\widetilde{v}} = {\rm LSTM}_{\widetilde{v}}^{R} (\mathbf{z}^{\widetilde{v}}_i, \mathbf{h}_{i-1}^{\widetilde{v}}, \mathbf{c}_{i-1}^{\widetilde{v}}).
\end{equation}
Here we regard the yielded hidden state $\mathbf{h}_i^{\widetilde{v}}$ as the reconstruction of the video feature $\mathbf{v}_i$.

For the syntax reconstructor, the corresponding hidden state $\mathbf{h}_i^{\widetilde{s}}$ is used to predict the syntax token $s_i$ in the exemplar syntactic sequence $S$. With the introduced two reconstructors, the aforementioned encoders are encouraged to embed more useful semantic and syntactic information, and the generator is encouraged to absorb more valid semantic and syntactic information for the exemplar-based video captioning task.

\subsection{Training and Inference}
\noindent \textbf{Training.} Formally, the proposed  architecture is trained by minimizing three loss terms defined in Eq.~(\ref{eq_all_loss}), which involves the caption generation loss, the video reconstruction loss, and the syntax reconstruction loss:
\begin{equation}
\small
\setlength{\abovedisplayskip}{3pt}
\setlength{\belowdisplayskip}{3pt}
\begin{split}
\mathcal{L}(\theta,& \theta_{rec}) = \sum_{(V,S) \in \Gamma_{train} } (
\underbrace{- \alpha{\rm log} P(W|V,S;\theta)}_{\text{caption generation}} \\
& \underbrace{+\lambda \mathcal{L}_{rec}^v(\mathbf{V},\mathbf{H}^{\widetilde{v}};\theta,\theta_{rec}^v)}_{\text{ video reconstruction}} \underbrace{- \eta{\rm log} P(S|V,S;\theta,\theta_{rec}^s)}_{\text{ syntax reconstruction}}
).
\end{split}
\label{eq_all_loss}
\end{equation}
$\alpha$, $\lambda$, and $\eta$ are the hyper-parameters to balance the contributions of different loss terms.
The video feature reconstruction loss $\mathcal{L}_{rec}^v$ is calculated by averaging the Euclidean distances between the original and reconstructed video features:
\begin{equation}
\small
\setlength{\abovedisplayskip}{3pt}
\setlength{\belowdisplayskip}{3pt}
 \mathcal{L}_{rec}^v(\mathbf{V},\mathbf{H}^{\widetilde{v}}) = \frac{1}{m} \sum_{i=1}^m {\rm Euclidean}(\mathbf{v}_i,\mathbf{h}_{i}^{\widetilde{v}}).
\end{equation}
The caption generation and syntax reconstruction losses are realized by the typical negative log-likelihood loss adopted by most video captioning methods.

Please note that we do not have the corresponding video caption which shares the same sentence syntax as the given exemplar sentence. Therefore,  when  training  our model, the ground-truth caption accompanied with the video is also used as the exemplar sentence. The proposed SMCG, the syntax recostructor, and the video reconstructor are then trained together to generate the ground-truth caption, reproduce its syntactic information, and reconstruct the video features, respectively.



\noindent \textbf{Inference.} During inference, we take the video and one randomly sampled exemplar sentence as the inputs of our model, then forward them through the encoders and the caption generator, and finally generate the exemplar-based video caption.

\vspace{-0.1in}
\section{Experiments}
\subsection{Datasets and Exemplar Sentence Collection}
Since there is no available benchmark datasets for exempler-based video captioning, we augment two public video captioning datasets MSRVTT \cite{xu2016msr} and ActivityNet Captions \cite{krishna2017dense} by collecting  auxiliary exemplar sentences for each video in these datasets. We will first give a brief introduction of the two datasets and then present our exemplar sentence collection procedure.

\noindent \textbf{MSRVTT \cite{xu2016msr}}. The MSRVTT is a large-scale dataset for video captioning. In this paper, we use the initial version of MSRVTT, which contains 10k videos from 20 categories. Each video is annotated with 20 ground-truth captions, resulting in a total of 200K video-caption pairs. We use the public splits for training and testing, i.e., 6,513 videos for training, 497 for validation, and 2,990 for testing.

\noindent \textbf{ActivityNet Captions \cite{krishna2017dense}}. The ActivityNet Captions is originally exploited as the benchmark for dense video captioning. This dataset contains 20K videos in total, and for each video, the temporal segments and their paired caption sentences are annotated. Since we do not consider dense video captioning in this paper, we split the caption-annotated segments from the original videos, and perform video captioning on them. In this way, 54,926 segment-caption pairs are collected, in which 37,421 segments split from the public training set are used for training, and 17,505 segments from the validation set are used for testing.



\noindent \textbf{Exemplar Sentence Collection.} To perform our exemplar-based video captioning, the exemplar sentences should have various syntactic structures. We find that previous work  \cite{Feng2018Unsupervised} crawled sentence descriptions from online stock photography website Shutterstock for unsupervised image captioning. The image descriptions they crawled are written by the image composer and free of syntactic constraints, which are suitable for the exemplar sentences we need. We download their collected 2,322,628 image descriptions, and filter the descriptions that are too short ($\textless$ 8 words) or too long ($\textgreater$ 30 words), yielding a total of 761,582 exemplar sentences. For each video/segment in MSRVTT and ActivityNet Captions, we randomly sample 20 collected descriptions as its exemplar sentences for captioning.

\subsection{Evaluation Metrics}
Different from the conventional video captioning, our exemplar-based video captions should describe the semantic contents of videos while imitate the syntactic structures of the exemplar sentences. Therefore, the evaluation of the exemplar-based captions should also be established in two aspects, i.e., semantic aspect and syntactic aspect.



For semantic evaluation, we firstly remove the stop words from each sentence and compute the averaged GloVe \cite{pennington2014glove} embeddings of the remaining words as its sentence embeddings. In the embedding space, the cosine similarity (COS) between the predicted exemplar-based caption and the original ground-truth video captions within the dataset is used to evaluate their semantic similarity. In addition, we also report the typical video captioning metrics BLEU@4, METEOR, ROUGE-L, and CIDEr as a reference.

For syntactic evaluation, we first parse the exemplar sentence and its corresponding predicted caption using Stanford CoreNLP \cite{Manning2014The}, and then compute the syntactic Tree Edit Distance (TED) \cite{zhang1989simple} between their constituency parse trees after removing word tokens. The smaller TED value means higher syntactic similarity. Although there are 20 exemplar sentences for each video, not each exemplar syntactic structure is suitable for describing the video. Therefore, we take the averaged TED between the 20 (exemplar, prediction) pairs as the result for each video, and diminish the impact of some unreliable exemplar sentences.

\subsection{Implementation Details}
We use the public Stanford parser \cite{Manning2014The} toolkit to process each exemplar sentence, and get the off-the-shelf constituency parse tree. Also note that the stanford parser will not be tuned in our overall architecture, but
other available sentence parsing networks \cite{shen2018straight,kitaev2018constituency} can also be applied in our work, and be jointly trained in an end-to-end fashion.


To extract video features, we feed the static video frames to the Inception-Resnet-v2 network \cite{Szegedy2016Inception} pretrained on the ILSVRC-2012-CLS image classification dataset \cite{Russakovsky2015ImageNet}, thus yielding a 1,536 dimensional feature vector for each frame. Considering both the video length distribution and model memory footprint, we take evenly spaced 30 and 100 features to represent videos in the MSRVTT and ActivityNet Captions datasets, respectively. Shorter videos of less than 30 or 100 features are padded with zero vectors.

In our model, we set the word embedding size and all the LSTMs' hidden size as 512, and the embedding size of the syntax token is set as 256. The trade-off parameters $\alpha$, $\lambda$, and $\eta$ of our model to balance different loss terms are set as 1.0, 1.0, and 4.0, respectively. For training, the model is optimized by the Adam \cite{KingmaAdam} optimizer.

\begin{table}[!t]
\footnotesize
\setlength{\abovecaptionskip}{0.cm}
\setlength{\belowcaptionskip}{-0.cm}
  \centering
  \caption{\footnotesize Performance comparisons on the MSRVTT dataset (*).}
  \label{tab:msrvtt}
    \begin{tabular}{p{0.25\columnwidth}p{0.07\columnwidth}p{0.07\columnwidth}p{0.07\columnwidth}p{0.07\columnwidth}p{0.07\columnwidth}p{0.07\columnwidth}}
    \toprule
    Method
    &TED$\downarrow$
    &COS$\uparrow$
    &B@4$\uparrow$
    &R$\uparrow$
    &M$\uparrow$
    &C$\uparrow$ \cr
    \midrule

   Exemplar-based
    &0.00 &52.38  &0.39
    &16.65 &6.02 &0.26 \cr

    Caption-based
    &15.91 &69.49  &37.20
    &58.64 &26.39 &40.24 \cr

    Concate-based
    &3.76 &66.18 &4.54
    &29.69 &15.24 &8.92 \cr
    \midrule
    SMCG
    &3.46 &67.02 &4.89
    &29.62 &15.64 &10.67 \cr

    SMCG+VideoRec
    &3.25 &67.26 &5.07
    &29.92 &15.95 &12.07 \cr

    SMCG+SyntaxRec
    &3.19 & 67.15 &4.65
    &29.45 &15.49 &10.50 \cr

    \textbf{SMCG+AllRec}
    &\textbf{3.12} &\textbf{67.97} &\textbf{5.01}
    &\textbf{30.41} &\textbf{16.25} &\textbf{12.22} \cr

    \midrule

    SMCG(BN)+AllRec
    &23.09 &61.22 &5.00
    &28.31 &13.92 &1.92 \cr

    SMCG(POS)+AllRec
    &6.33 & 67.14 &5.08
    &29.92 &15.83 &12.03 \cr

    \bottomrule
    \end{tabular}
  \begin{tablenotes}
        \scriptsize
        \item [1] *: TED is reported with absolute value, while other metrics are reported in percentage \% values. $\downarrow$ means smaller values are better, and $\uparrow$ is on the contrary.
      \end{tablenotes}
\end{table}

\subsection{Performance Comparison}

Since there is no existing work for exemplar-based video captioning, we compare our proposed method with three baseline methods as follows. ``\textbf{Exemplar-based}'' method directly outputs the exemplar sentence as the predicted caption, while does not consider the video contents. ``\textbf{Caption-based}'' method applies the general sequence-to-sequence video captioning model \cite{yao2015describing}, while does not consider the exemplar syntax information. ``\textbf{Concate-based}'' method takes the main architecture of ``Caption-based'' method, but also leverages another encoder to sequentially encode syntax information as we did. Then, the video context features and syntactic context features are directly concatenated as the input to the caption decoder. For comparison, our proposed full model is denoted by SMCG+AllRec.

As shown in the upper part of Table \ref{tab:msrvtt} and Table \ref{tab:anet}, Exemplar-based method gets a TED score 0.0, which accords with that it simply takes the given exemplar sentence as the predicted caption. However, since it ignores video contents, the semantic metrics such as COS, BLEU@4, ROUGE-L, METEOR, and CIDEr are much lower than other methods. In contrast, Caption-based method just considers describing video contents while neglects syntactic requirements of the predicted caption, making higher semantic metric scores but much larger TED values.


For exemplar-based video captioning, the small TED scores of the Concate-based method verify that its predicted captions have indeed imitated the syntactic structures of the given exemplar sentences. Since general caption evaluation metrics like ROUGE take n-gram similarity between sentences into consideration, they will be inevitably influenced by the exemplar syntactic constraints. As such, compared with the Caption-based method, the general captioning metric values of the Concate-based method significantly decrease. Sentence cosine similarity only considers word semantics in sentences, so its values are more stable. However, directly concatenating the syntactic information with video features in the captioning procedure will cause disturbances in describing semantic video contents, thus yielding smaller COS values of the Concate-based method than the Caption-based method.


Compared with the Concate-based method, our proposed model SMCG+AllRec steps further in generating syntax customized and semantic preserved video captions. As shown in Table \ref{tab:msrvtt} and Table \ref{tab:anet}, SMCG+AllRec achieves smaller TED values and higher semantic metric scores on both two datasets. On one hand, our method leverages the syntactic information to meticulously modulate the LSTM updating procedure while does not directly alter the semantic inputs of the decoder. On the other hand, the reconstruction of video features and syntax token sequences can further help preserve the video semantic contents and exemplar syntactic structures in the predicted captions, respectively. By incorporating both of the above aspects, our proposed SMCG+AllRec consistently achieves the best results.

\begin{table}[!t]
\footnotesize
\setlength{\abovecaptionskip}{0.cm}
\setlength{\belowcaptionskip}{-0.cm}
  \centering
  \caption{\footnotesize Performance comparisons on the ActivityNet Captions dataset.}
  \label{tab:anet}
     \begin{tabular}{p{0.25\columnwidth}p{0.07\columnwidth}p{0.07\columnwidth}p{0.07\columnwidth}p{0.07\columnwidth}p{0.07\columnwidth}p{0.07\columnwidth}}
    \toprule
    Method
    &TED$\downarrow$
    &COS$\uparrow$
    &B@4$\uparrow$
    &R$\uparrow$
    &M$\uparrow$
    &C$\uparrow$ \cr
    \midrule
    Exemplar-based
    &0.00 &57.84 &0.07
    &6.74  &2.57 &0.98 \cr

    Caption-based
    &20.12 &73.17 &3.36
    &20.44 &8.93 &23.92 \cr

    Concate-based
    &3.71 &64.27 &0.49
    &10.90 &4.86 &7.83 \cr

    \midrule
    SMCG
    &3.41 &64.91 &0.54
    &10.91 &4.86 &8.95 \cr

    SMCG+VideoRec
    &3.64 &66.46 &0.54
    &10.91 &4.98 &9.14 \cr

    SMCG+SyntaxRec
    &3.30 &67.50 &0.53
    &10.88 &4.92 &8.06 \cr

    \textbf{SMCG+AllRec}
    &\textbf{3.05} &\textbf{68.23} &\textbf{0.53}
    &\textbf{11.11} &\textbf{5.11} &\textbf{9.31} \cr

    \midrule

    SMCG(BN)+AllRec
    &8.48 &61.11 &0.39
    &9.36 &3.87 &3.32 \cr

    SMCG(POS)+AllRec
    &6.02 &68.17 &0.54
    &11.10 &5.09 &9.29 \cr

    \bottomrule
    \end{tabular}
\end{table}



\subsection{Ablation Studies}
To better demonstrate the effectiveness of our model design, we perform several ablation studies as follows:

\noindent \textbf{SMCG:} We only keep the semantic and syntactic encoders, and the syntax modulated caption generator in the proposed architecture, with the two reconstructors removed.

\noindent \textbf{SMCG+VideoRec:} Based on SMCG, the video reconstructor is included to help the model training.

\noindent \textbf{SMCG+SyntaxRec:} Based on SMCG, the syntax reconstructor is included to help the model training.

\noindent \textbf{SMCG+AllRec:} Our proposed full model.

\noindent \textbf{SMCG(BN)+AllRec:} In this setting, the feature normalization procedure in our proposed syntactic modulation operation is replaced by batch normalization, instead of the originally used layer normalization.

\noindent \textbf{SMCG(POS)+AllRec:} In this setting, the sentence syntactic information is represented with the POS sequence of words as used in the previous work \cite{wang2019controllable}, instead of the constituency parse tree in our model.

\begin{figure*}[!t]
\centering
\setlength{\abovecaptionskip}{0.cm}
\setlength{\belowcaptionskip}{-0.cm}
\includegraphics[width=1.0\textwidth]{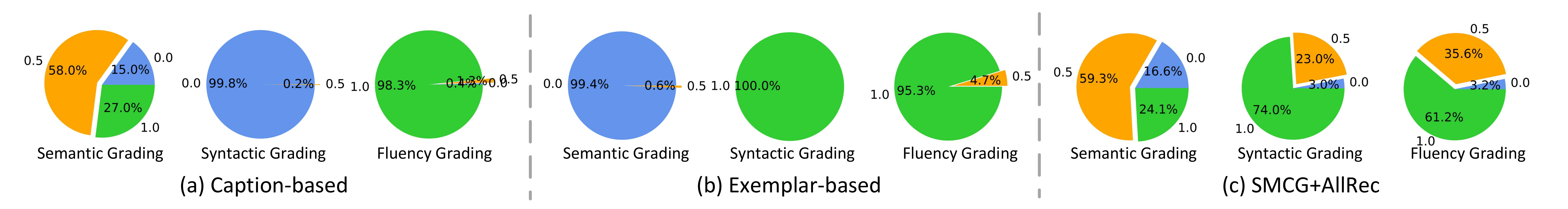}
\caption{Subjective evaluations on the MSRVTT datasets.}
\label{fig:pie_chart}
\end{figure*}


The ablation experimental results are also shown in  Table \ref{tab:msrvtt} and Table \ref{tab:anet}.  By incorporating video reconstruction in SMCG+VideoRec, the semantic metric values improve on both the datasets, which demonstrates the benefits of the video feature reconstruction for preserving video semantics. In addition, incorporating syntax reconstruction in SMCG+SyntaxRec yields smaller TED scores than SMCG, which indicates the syntax reconstruction can further help the syntax customization. Interestingly, we also find that incorporating video reconstruction makes smaller TED scores on the MSRVTT dataset, and incorporating syntax reconstruction yields higher COS scores on the ActivityNet captions dataset. Such observations show that a better understanding of video contents may help the model organize sentence syntaxes, and incorporating valid syntax constraints may drive the model to present video contents better. Combining both of the two reconstructions together, our full model SMCG+AllRec yields further improvements in both TED and COS scores.


By changing the feature normalization method from channel-wise layer normalization to batch normalization, we find that the performances of SMCG(BN)+AllRec drop drastically compared to SMCG+AllRec. Although batch normalization has made achievements in CNN and RNN architectures \cite{de2017modulating,dumoulin2016learned,cooijmans2016recurrent}, such a normalization scheme is not very suitable for our proposed SMCG. Since one batch in our modeling training procedure contains several (video, exemplar syntax) pairs, the normalization across the whole batch will make disturbances between different training instances and degrade the model performance.

Moreover, we also find that replacing sentence syntactic representation from the constituency parse tree to the POS sequence makes the TED of SMCG(POS)+AllRec increase a lot compared to SMCG+AllRec. Such results indicate that constituency parses represent sentence syntaxes better than POS sequences, and are more suitable for our exemplar-based syntax customized video captioning.


\vspace{-0.08in}
\subsection{Subjective Evaluation}

\begin{figure*}[!t]
\centering
\includegraphics[width=1.0\textwidth]{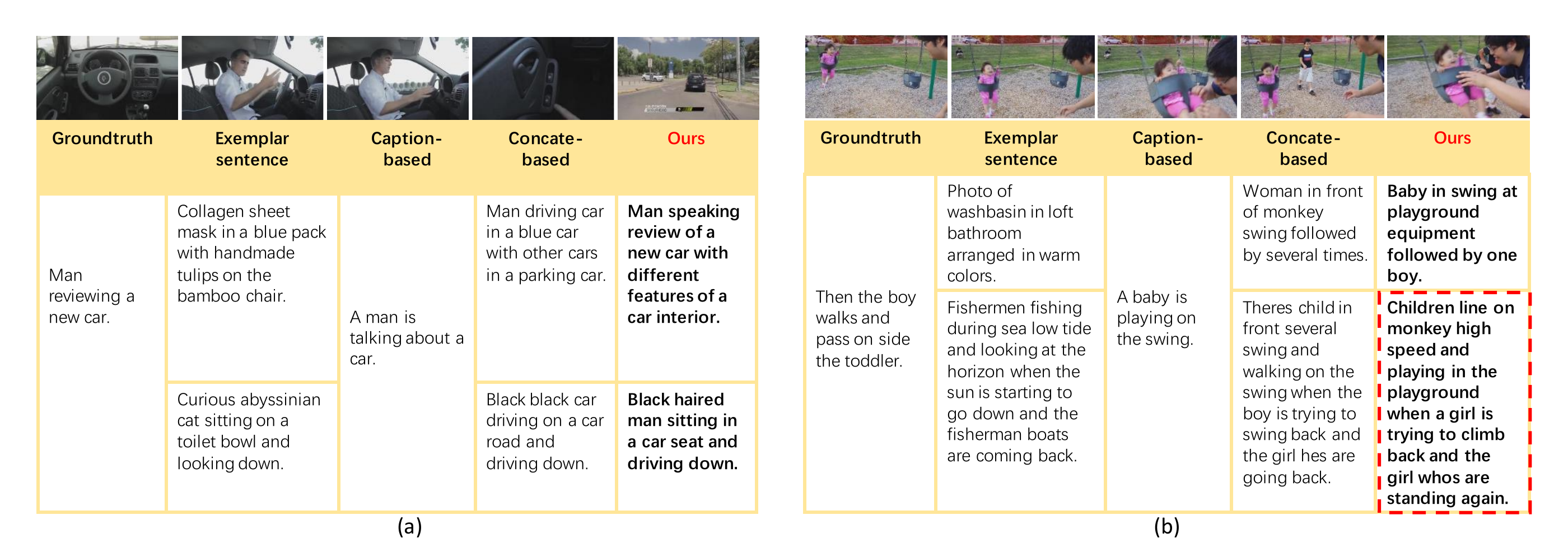}
\caption{ \small Qualitative results for some exemplar-based video captioning cases. In the first column of each case, we provide the ground-truth video captions. In the second column, we give two exemplar sentences for syntactic constraints. The predicted captions by the Caption-based method, the Concate-based method, and our method are presented in the third, fourth, and fifth columns, respectively. Case (a) is from the MSRVTT dataset, and case (b) is from the ActivityNet Captions dataset.}
\label{fig:quality}
\end{figure*}

Although COS and TED metrics can evaluate the quality of the predicted sentences from semantic and syntactic aspects, it is more reliable to conduct a human study to judge whether our method can produce reasonable semantic-preserved and syntax-controlled captions. Therefore, we invite 18 evaluators (9 females and 9 males) from different education backgrounds to perform the subjective evaluation. For both of the two datasets, we randomly choose 300 samples, and each of them contains one video and one exemplar sentence. The evaluators are required to watch the video first, and then evaluate the predicted captions from the semantic, syntactic and fluency aspects by grading it on three scales (0, 0.5, 1). Grade 1 means that the predicted caption presents the video semantics well/follows the exemplar syntax well/has good linguistic fluency, 0.5 means not bad, and 0 means poor. Three groups of captions generated by the Caption-based method, the Exemplar-based method, and our proposed SMCG+AllRec method are graded by these evaluators for comparisons. The ratios of different grades on the MSRVTT dataset are illustrated in Figure \ref{fig:pie_chart}, while the results on the ActivityNet Captions dataset are provided in our supplemental material.

Specifically, we can observe that for the semantic subjective evaluation, our proposed SMCG+AllRec method achieves comparable results with the Caption-based baseline, and there are 83.4\% and 85\% records graded equal or larger than 0.5 for these two methods, respectively. It shows that both of these two methods can generate captions describing the video semantic contents well. However, on the conventional captioning metrics (e.g., B@4) as shown in Table \ref{tab:msrvtt} and Table \ref{tab:anet}, our method and the Caption-based baseline have significant performance gap. The different evaluation results indicate that the conventional captioning metrics, which are influenced by sentence syntaxes/n-gram characteristics, indeed have limitations on evaluating the exemplar-based  video captioning task. For the syntactic evaluation, the Caption-based method gets 99.8\% records graded at 0.0, which shows that this method cannot make the predicted captions follow the exemplar sentence syntaxes. In contrast, the grade 0.0 only occupies 3\% syntactic evaluation records for our SMCG+AllRec method, which verifies that our generated captions can not only present the video semantics well, but also follow the exemplar syntaxes well. As for the Exemplar-based baseline, it gets (99.4\%,0.6\%,0\%) and (0\%,0\%,100\%) records on (0,0.5,1) grades for semantic and syntactic evaluation. It is evident that the given exemplar sentences are totally irrelevant to video semantics.

When evaluating the sentence fluency of the predicted captions, there are 98.3\%, 95.3\%, and 61.2\% records graded at 1.0 for the Caption-based method, the Exemplar-based method, and the proposed SMCG+AllRec method, respectively. The Caption-based method generates captions by fitting the simple and monotonous sentence syntaxes (like ``A is doing B'') of the training captions in the dataset. Such simple syntactic structures are easy to learn by the captioning model, therefore making the generated captions have good linguistic fluency. The provided exemplar sentences are human-edited sentences with more various and complex syntactic structures. Even a small fraction (4.7\%) of the exemplar sentences are not very easy to follow and are graded at 0.5 in fluency evaluation. By making the generated captions follow such complex exemplar syntaxes, our proposed SMCG+AllRec method can still generate fluent captions to a large extent with only 3.2\% records graded at 0.0. It shows that our proposed method can make the captions maintain good linguistic fluency in the meanwhile of imitating the exemplar syntaxes.


\newcommand{\tabincell}[2]{\begin{tabular}{@{}#1@{}}#2\end{tabular}}
\begin{table}[!t]
\setlength{\abovecaptionskip}{0.cm}
\setlength{\belowcaptionskip}{-0.cm}
  \centering
\small
\begin{adjustbox}{max width = \textwidth}
  \begin{threeparttable}
  \caption{ Captioning Diversity Evaluation.}
  \label{tab:diversity}
    \begin{tabular}{ccccc}
    \toprule
    \multirow{3}{*}{Method}&\multicolumn{2}{c}{MSRVTT}&\multicolumn{2}{c}{ActivityNet Captions}\cr
    \cmidrule(lr){2-3} \cmidrule(lr){4-5}
    &\tabincell{c}{LSA}
    &\tabincell{c}{Self-CIDEr}
    &\tabincell{c}{LSA}
    &\tabincell{c}{Self-CIDEr} \cr
    \midrule
    Caption-based &0.0 &0.0 &0.0 &0.0  \cr
    Exemplar-based &0.7437 &0.9737 &0.7431 &0.9741 \cr
    SMCG+AllRec &0.5896 &0.8628 &0.5862 &0.8748 \cr
    \bottomrule
    \end{tabular}
    \end{threeparttable}
 \end{adjustbox}
\end{table}

\subsection{Captioning Diversity Evaluation}

Given different exemplar sentences (we have 20 exemplar sentences for each video in our collected dataset) and one single video, our SMCG+AllRec can generate different captions describing the semantic contents of the video, and thus it can be regarded as diverse video captioning. To evaluate the diversity of the generated captions, we follow the evaluation metrics introduced in the work \cite{wang2019diversity}, and report the LSA and Self-CIDEr based diversities of different methods in Table \ref{tab:diversity}. Since the Caption-based method can only generate one single caption for one video no matter what exemplar sentence is provided, the diversity values of its generated captions are always 0.0. The Exemplar-based method directly outputs the exemplar sentences as the caption prediction results, and the provided 20 exemplar sentences in our experiments are different from each other. Therefore, the diversity values of the Exemplar-based method can be seen as the upper bounds for the syntax-controllable video captioning. For our SMCG+AllRec method, it generated captions achieve high Self-CIDEr diversities 0.8628 and 0.8748 on the MSRVTT and ActivityNet Captions dataset, respectively. Meanwhile, the LSA based diversities are also comparable to other diverse video captioning methods as reported in \cite{wang2019diversity}. Therefore, the results verify the contribution of our method on strengthening the diversity of video captioning.

\subsection{Qualitative Results}

Finally, we show some qualitative results in Figure \ref{fig:quality} for the proposed exemplar-based video captioning. It can be observed that by providing different exemplar sentences, our method can generate different video captions with various syntactic structures, which enhances the diversity and expressiveness of the video descriptions compared to the Caption-based method in the third column. Meanwhile, our method can generate more accurate and fluent sentence descriptions than the Concate-based method, which indicates the effectiveness of our SMCG design in preserving video semantics and controlling sentence syntaxes.

In addition, we also provide a failure case of the exemplar-based video captioning, as shown in the red dotted box in Figure \ref{fig:quality}. The given exemplar sentence of this case is relatively long and complex, making our predicted video caption unfluent and confusing. It demonstrates that the exemplar sentence selection has a great influence on the final video captioning results. Different videos are suited to different descriptive manners, and we should choose appropriate exemplar sentences for them to match. However, establishing the matching relations between semantic video contents and syntactic sentence structures is a challenging problem, which is not the main focus of this paper, and will be explored in the future. For more qualitative results, please refer our supplemental material.

\section{Conclusion}
In this paper, we proposed a novel exemplar-based video captioning problem, which aims to generate one natural sentence describing the video content and meanwhile following the syntactic structure of the provided exemplar sentence. A novel syntax modulated caption generator (SMCG) was proposed to leverage encoded syntactic information to modulate the LSTM gates and cells for decoding each word, therefore controlling the predicted sentence syntax. Moreover, the video and syntax reconstructors are further introduced to help preserve video semantics in predicted captions and ensure the syntax customization. Extensive experiments demonstrate that our proposed method can yield different video descriptions of various syntax structures with respect to different exemplar sentences, hence enhancing the diversity of video captions.

\section{Acknowledgments}
This work was supported by National Natural Science Foundation of China Major Project (No. U1611461) and National Key R\&D Program of China under Grand No. 2018AAA0102000.

\balance
\bibliographystyle{ACM-Reference-Format}
\bibliography{sample-base}

\end{document}